\documentclass[conference]{IEEEtran}
\IEEEoverridecommandlockouts
\usepackage{cite}
\usepackage{amsmath,amssymb,amsfonts}
\usepackage{algorithmic}
\usepackage{graphicx}
\usepackage{textcomp}
\usepackage{xcolor}
\usepackage{todonotes}
\usepackage[hidelinks, 			
breaklinks=true		
]{hyperref}	
\def\BibTeX{{\rm B\kern-.05em{\sc i\kern-.025em b}\kern-.08em
		T\kern-.1667em\lower.7ex\hbox{E}\kern-.125emX}}
\begin{document}
	
\title{%
  Content-Aware Depth-Adaptive Image Restoration\\
}
	
	\author{\IEEEauthorblockN{Tom Richard Vargis, Siavash Ghiasvand}
		\IEEEauthorblockA{Center for Interdisciplinary Digital Sciences (CIDS), Technische Universität Dresden, Germany}
        \IEEEauthorblockA{Center for Scalable Data Analytics and Artificial Intelligence (ScaDS.AI) Dresden/Leipzig. Germany\\
        \{tom\_richard.vargis , siavash.ghiasvand\}@tu-dresden.de}
	}
	
	\maketitle
	
	


\section*{}
\label{sec-motivation}
The pursuit of enhancing visual clarity, detail, and overall aesthetic appeal in image restoration has evolved with the emergence of deep learning models. While existing models like Midjourney, Stable-Diffusion, and DALL-E 2 simplify the restoration process with prompts, they lack fine-grained user control. Furthermore, the generated output is highly non-deterministic. This research introduces a modular pipeline designed to empower users with comprehensive control over image restoration, demanding minimal skills and time. The pipeline independently recognizes and restores individual objects within the input image, allowing for precise control over their categorical restoration, placement, and layering. Its modular design facilitates easy substitution of models, enhancing adaptability to specific object domains such as medical image restoration.
To facilitate the reproducibility of this work, the source code and sample data are publicly available~\cite{tom2023}.

Several research efforts have aimed to enhance image restoration by introducing additional functionalities. Despite the advantages of text-based approaches highlighted in \cite{bkawar23}, and \cite{dxie23}, they fall short in providing fine-grained user control, especially with severely distorted images. Although the methodologies proposed in \cite{alugmayr22}, and \cite{clee20} involve direct user participation in restoration, such as controlling denoising and deblurring or modifying masks for personalized preferences, they typically adopt a coarser granular approach, treating the entire image as a single entity. Existing content-aware methods like \cite{tyu23}, \cite{mweigert18}, and \cite{jjiang23} often focus on individual issues, lacking a single method encompassing all benefits. 
The proposed pipeline integrates functionalities with unique content awareness, providing users complete control over image edits with user-friendly accessibility. The pipeline's adaptability suggests potential extensions to other specialized image domains like in~\cite{mweigert18}.
Moreover, the suggested content-aware modular pipeline, provides a deterministic restoration method even in the face of underlying models' unpredictable behavior, thereby ensuring that each produced outcome is explainable and reproducible.

\section*{}
\label{data-and-pipeline}
The proposed pipeline comprises distinct stages, each processed separately and replaceable based on user preferences. Users can configure the pipeline using a simple dictionary file, tailoring it to their specific needs. The modular framework integrates existing deep learning models, leveraging their strengths to enhance the overall performance of the pipeline. This approach stands in contrast to traditional methods that create models from scratch, providing a more flexible and efficient restoration process. The pipeline's dynamic stages adapt to user-defined configurations, offering fine-grained control over the image restoration. A directory-based approach facilitates image-in, and image-out processing at each stage, ensuring a seamless flow of information. Users can view and alter the models for any stage, adding a layer of flexibility to the automated workflow.\\
The pipeline isolates the contents of the image using a combination of object detection [YOLOv3u, YOLOv8n]\footnote{\url{https://github.com/ultralytics/ultralytics}} and background removal [DeepLabv3\footnote{\url{https://pytorch.org/hub/pytorch_vision_deeplabv3_resnet101/}}, YOLOv8-seg] algorithms (Path 1) or by alternatively using an instance segmentation [YOLOv8-seg] algorithm (Path2) before proceeding to image restoration [runwayml/StableDiffusion-inpainting]\footnote{\url{https://github.com/runwayml/stable-diffusion}}. The models independently restore each identified object and its background and reposition them to generate the final output. It is important to highlight that the mentioned models are solely utilized as a proof of concept, and any other model relevant to the image content can be employed.

In the inpainting image restoration step, a mask is required to selectively restore essential regions of the instance rather than the entire entity. An additional finer granularity control is achieved through image tuning that allows the user to modify individual objects based on a prompt and layered scene customization that facilitates depth adaptive remodeling of the image. This component of the pipeline allows users to seamlessly remove objects from the image and restructure them by adjusting their positions in the x, y, and z directions within the scene. These two additional steps provide extra control and may not necessarily be integral components of the main pipeline branch. The models selected for each component are pre-trained and chosen based on their efficiency in handling the COCO animal dataset. 
A wrapper code facilitates the seamless integration of the models into the pipeline, ensuring a smooth execution of the restoration process. Additional functionalities, such as padding the cropped instances and morphological operations on masks, enhance the overall regeneration process.\\

The evaluation employed a dataset featuring 100 images of animals\footnote{Unplash: Free images.\url{https://unsplash.com/}} namely cats, dogs, elephants, horses, and zebras purposefully distorted especially in facial regions. A key objective was to assess the pipeline's performance against a direct regeneration method by utilizing the same inpainting model\footnote{runwayml/stable-diffusion-inpainting} with identical internal parameters for both approaches.
The evaluation encompasses both objective and subjective measures, including class probability analysis, comparative scatter plots of confidence scores, average gains in confidence scores, mean variation in confidence scores, and a subjective assessment of the restored images. To ensure a comprehensive evaluation, the ground truth was established using the YOLOv3u model to detect classes and their respective probabilities in the original images.\\
Objective comparisons involve a detailed class probability analysis, showcasing the impact of manual distortions on confidence scores with an average drop of 5.3\% across the entire dataset and its improvement with different restoration methods. Scatter plots illustrate the relationship between ground truth scores and confidence scores obtained from different restoration methods. The analysis extends to average gains in confidence scores, revealing notable improvements in categories like zebras, horses, and elephants through the pipeline. These specific animal categories experienced a 6.6\% boost in confidence scores when restored via the pipeline, surpassing the 5.16\% improvement observed with the direct method. Conversely, confidence scores for cat and dog categories declined due to misclassification during the initial stage of the pipeline, presenting a counterproductive outcome. A mean variation in confidence scores provides insights into the error rates, with the stable diffusion method consistently maintaining low error rates across all categories. While the pipeline restorations outperformed the ground truth for zebras, horses, and elephants, it still suffers from the misclassification impacting the cat and dog categories leading to increased error rates.\\
Subjective comparisons involve a visual assessment of the restored images, in terms of well-defined features and image quality. The study explores the optimal pathway within the pipeline by considering factors such as complexity, consistency, and time efficiency. Path 2 emerges as the preferred choice due to its simplified structure, combining object detection and background removal into a single-step instance segmentation. The pipeline approach is seen to be far superior in this evaluation with restoration resulting in more defined features for these animal categories, while the direct approach merely smudges the distorted region.\\ 
In conclusion, the proposed pipeline proves to be a compelling proof of concept for image restoration, offering users unprecedented control and flexibility. The analysis underscores the pipeline's superiority over direct regeneration methods in terms of image quality, albeit contingent on the quality of employed models. However, it was also noticed that the pipeline is only as good as the models used and that issues like false detection can impact the performance significantly. The study sets the stage for future advancements, including parallelization of the sequential restoration process, automated model selection based on image characteristics and the integration of super-resolution and retouching models for enhanced capabilities.

\section*{}
\subsubsection*{Acknowledgements}
\label{acknowledgements}
This work was supported by the German Federal Ministry of Education and Research (BMBF, SCADS22B) and the Saxon State Ministry for Science, Culture and Tourism (SMWK) by funding the competence center for Big Data and AI "ScaDS.AI Dresden/Leipzig“.
The authors gratefully acknowledge the GWK support for funding this project by providing computing time through the Center for Information Services and HPC (ZIH) at TU Dresden.

%
%
%
%


\end{document}